\documentclass[letterpaper, 10 pt, conference]{ieeeconf} 
\usepackage{gensymb}
\usepackage{hyperref}
\usepackage{tabularx}
\usepackage{tabu}
\usepackage{graphicx}
\usepackage{caption}
\usepackage{subfigure}
\usepackage{color}
\usepackage{multirow}
\usepackage{amsmath}
\usepackage{array}
\usepackage{pifont}
\definecolor{pink_abhi}{rgb}{0.858, 0.188, 0.478}

\IEEEoverridecommandlockouts                             
\title{\LARGE \bf
The H3D Dataset for Full-Surround 3D Multi-Object Detection and Tracking in Crowded Urban Scenes
}
\author{Abhishek Patil, Srikanth Malla, Haiming Gang and Yi-Ting Chen
\thanks{The authors are with Honda Research Institute, 375 Ravendale Dr, Suite B,
Mountain View, CA, 94043, USA.
{\tt\small \{apatil,smalla,hgang,ychen\}@honda-ri.com}}
}
\begin{document}
\maketitle
\thispagestyle{empty}
\pagestyle{empty}

\begin{abstract}
%
3D multi-object detection and tracking are crucial for traffic scene understanding. 
However, the community pays less attention to these areas due to the lack of a standardized benchmark dataset to advance the field.
Moreover, existing datasets (e.g., KITTI~\cite{kitti}) do not provide sufficient data and labels to tackle challenging scenes where highly interactive and occluded traffic participants are present.    
To address the issues, we present the Honda Research Institute 3D Dataset (H3D), a large-scale full-surround 3D multi-object detection and tracking dataset collected using a 3D LiDAR scanner.
H3D comprises of 160 crowded and highly interactive traffic scenes with a total of 1 million labeled instances in 27,721 frames.
With unique dataset size, rich annotations, and complex scenes, H3D is gathered to stimulate research on full-surround 3D multi-object detection and tracking. 
To effectively and efficiently annotate a large-scale 3D point cloud dataset, we propose a labeling methodology to speed up the overall annotation cycle.
A standardized benchmark is created to evaluate full-surround 3D multi-object detection and tracking algorithms.
3D object detection and tracking algorithms are trained and tested on H3D.
Finally, sources of errors are discussed for the development of future algorithms.
\end{abstract}

\section{INTRODUCTION}
Multi-object detection and tracking are two essential tasks for traffic scene understanding. 
The field has been significantly boosted by recent advances of deep learning algorithms~\cite{RenNIPS2015,HeCVPR2016,zhao2017pspnet,HuangDensenetCVPR2017,HeICCV2017} and an increasing number of datasets~\cite{imagenet_cvpr09,LinECCV2014_COCO,Cordts2016Cityscapes,CNeuholdICCV2017,HuangApolloCVPR2018}. 
While tremendous progress has been made in 2D traffic scene understanding, it still suffers from the fundamental limitations in the sensing capability and lack of 3D information.
Recently, with the emerging technology of 3D range scanners, the range sensor directly measure 3D distances by illuminating the environment with pulsed laser light. 
It enables a wide range of robotic applications in the 3D world. 
While 3D scene understanding is important for these applications, relatively small efforts~\cite{kitti,MaddernIJRR2016,semanticnet,JeongICRA2018} have been attempted in comparison to its 2D counterpart.
%

The Oxford RobotCar dataset~\cite{MaddernIJRR2016} was proposed to address the challenges of robust localization and mapping under significantly different weather and lighting conditions.
Recently, Jeong et al.~\cite{JeongICRA2018} introduced a complex urban LiDAR dataset collected in metropolitan areas, large building complexes, and underground parking lots.
However, these datasets mainly focus on Simultaneous Localization and Mapping (SLAM).

\begin{figure}[t]
    \centering
    \includegraphics[width=0.5\textwidth]{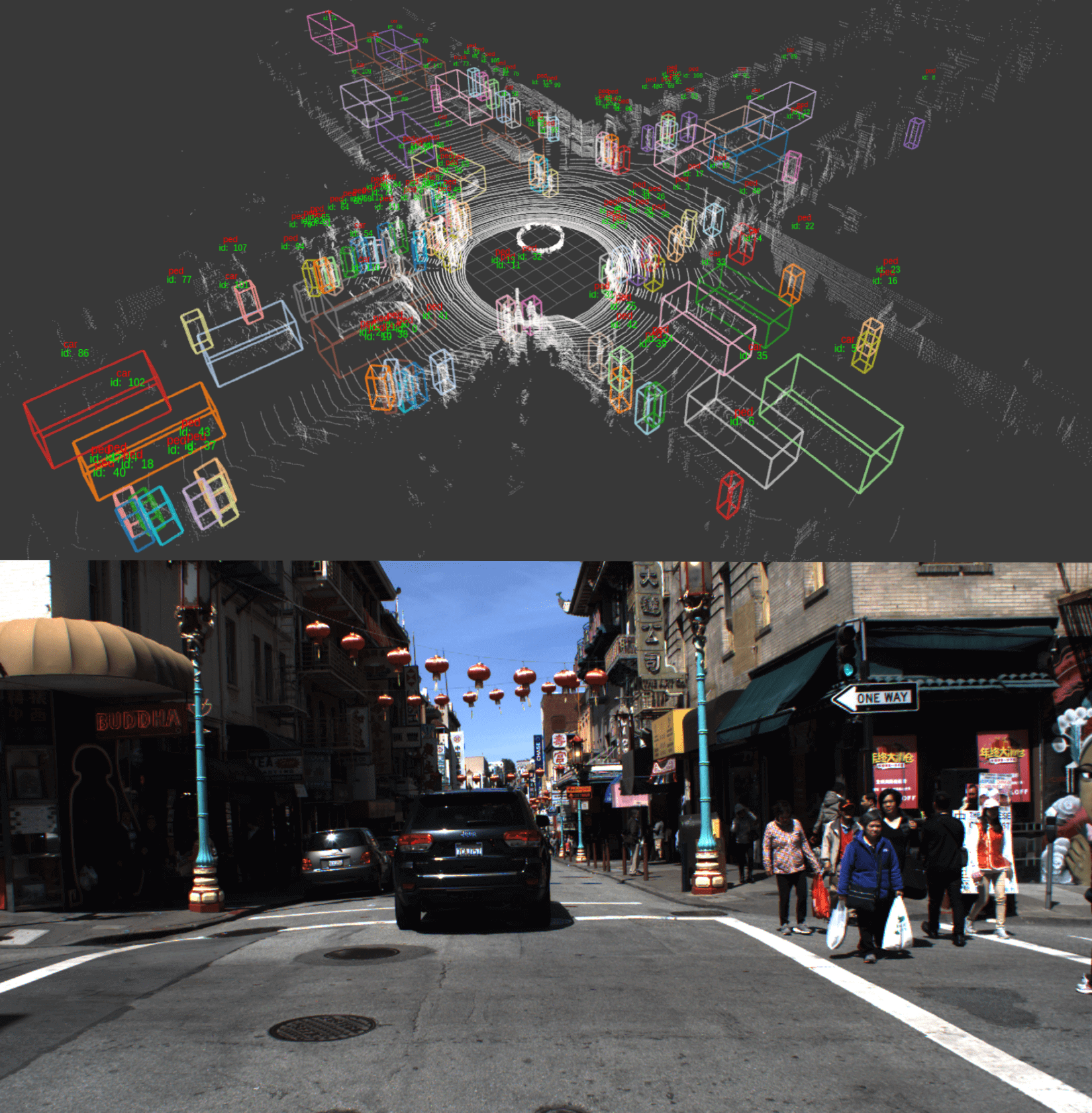}
    \caption{The Honda Research Institute 3D Dataset (H3D) for full-surround 3D multi-object detection and tracking in crowded urban scenes.}
    \label{3d_boxes}
\end{figure}

\begin{figure*}[h!]
    \centering
    \includegraphics[width=\textwidth]{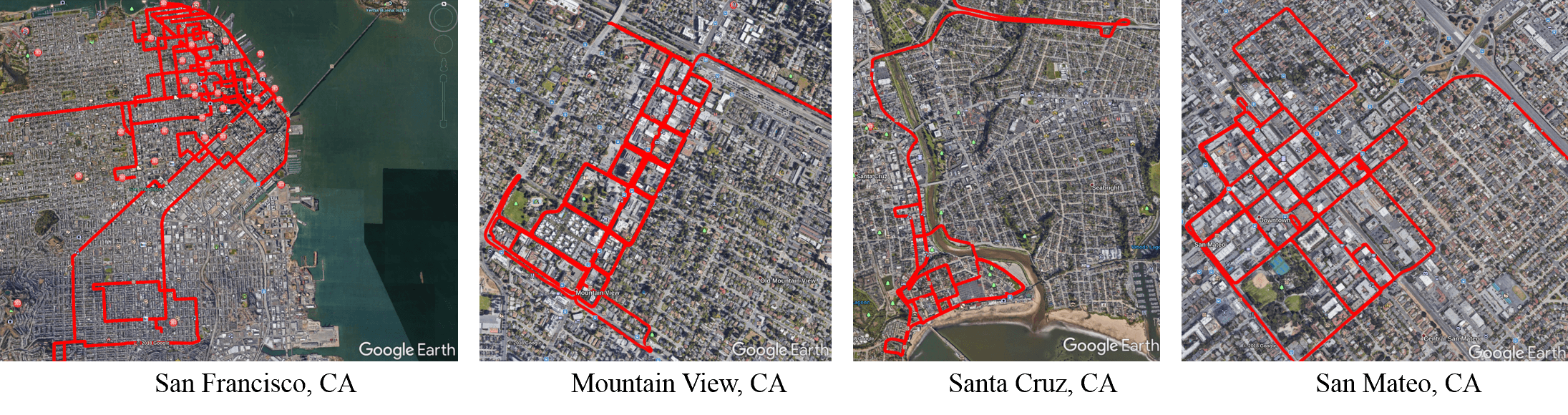}
    \caption{Geographical distribution of H3D.}
    \label{fig:data_region}
\end{figure*}

Among the existing attempts, KITTI dataset~\cite{kitti} enables various scene understanding tasks including 3D object detection and tracking.
Specifically, it comprises of more than 200k manually labeled 3D objects captured in cluttered scenes.
However, KITTI dataset is insufficient to advance the future development of 3D multi-object detection and tracking for the following reasons.
First, the 3D object annotations are only labeled in the frontal view that limits the applications required full-surround reasoning.
Second, KITTI dataset has relatively simple scene complexity without extensive data from crowded urban scenes, e.g., metropolitan areas where highly interacting and occluding traffic participants are present.
Third, the richness of existing labels in KITTI dataset is inadequate for deep learning algorithms to learn diverse appearances from data.
Fourth, KITTI dataset does not have a standardized evaluation for full-surround multi-object detection and tracking in 3D.  

To address the aforementioned issues, H3D is designed and collected with the explicit goal of stimulating research on full-surround 3D multi-object detection and tracking in crowded urban scenes. 
The H3D is gathered from HDD dataset\footnote{https://usa.honda-ri.com/HDD}~\cite{RamanishkaCVPR2018}, a large scale naturalistic driving dataset collected in San Francisco Bay Area. 
Diverse, rich, and complex traffic scenes are selected in four major urban scenes as shown in Fig.~\ref{fig:data_region} to develop and evaluate 3D multi-object detection and tracking algorithms.
To annotate a large-scale dataset, we establish an effective and efficient labeling process to speed up the overall annotation cycle. The details will be discussed in Sec.~\ref{sec:data_annotation}. 

The contributions are summarized as follows. First, H3D is the first dataset for full-surround 3D multi-object detection and tracking in crowded urban scenes comprising of 1,071,302 3D bounding box labels of 8 common traffic participants.
Second, a labeling methodology is introduced to annotate large-scale 3D bounding boxes. 
Third, a standardized benchmark of full-surround 3D multi-object detection and tracking is established for future algorithm developments.
The dataset is available at \url{http://usa.honda-ri.com/H3D}.

\section{TRAFFIC SCENE DATASETS}
An increasing number of 2D scene understanding datasets~\cite{Cordts2016Cityscapes,CNeuholdICCV2017,HuangApolloCVPR2018} are proposed in recent years. 
In particular, these datasets aim to stimulate research on semantic segmentation for traffic scenes by providing high quality labels and scalable dataset generation methodologies.
The Cityscapes dataset provides 5000 images with high quality pixel-level annotations and additional 20,000 images with coarse annotations for methods that leverage large volumes of weakly-labeled data~\cite{Cordts2016Cityscapes}.
The Mapillary dataset~\cite{CNeuholdICCV2017} increase the size of pixel-level annotations to 20,000 images and the diversity of images by selecting images from all around the world.
While the two datasets provide high quality and high volume 2D annotations, they lack information from 3D to enable research on 3D object detection and tracking.

With a comparison to 2D scene understanding, relatively small efforts~\cite{oaklanddataset,semanticnet,forddataset} have been made in 3D due to the costs for installing a high quality 3D range scanner and difficulties in labeling annotations in point cloud on a large-scale.
Semantic3D.Net~\cite{semanticnet} and Oakland dataset~\cite{oaklanddataset} are two point cloud datasets that provide semantic labels for point cloud classification. 
The Ford Campus LiDAR Dataset~\cite{forddataset} consists of point cloud data collected in urban environments from multiple LiDAR devices. 
However, 3D bounding boxes and tracks of objects are not available to enable research on 3D detection and tracking of traffic participants.

It is non-trivial to manually label large-scale datasets. 
Huang et al.~\cite{HuangApolloCVPR2018} annotate large-scale semantic segmentation by projecting labeled semantic labels on survey-grade dense 3D points.
In the proposed labeling methodology, we leverage a similar idea by applying LiDAR SLAM to register multiple LiDAR scans to form a dense point cloud. 
In this case, static objects will only have to be labeled once instead of a frame-by-frame annotation.
This methodology significantly improves the overall labeling cycle.
More details will be discussed in Sec.~\ref{sec:data_annotation}.
\begin{figure*}[h!]
    \centering
    \includegraphics[width=\textwidth]{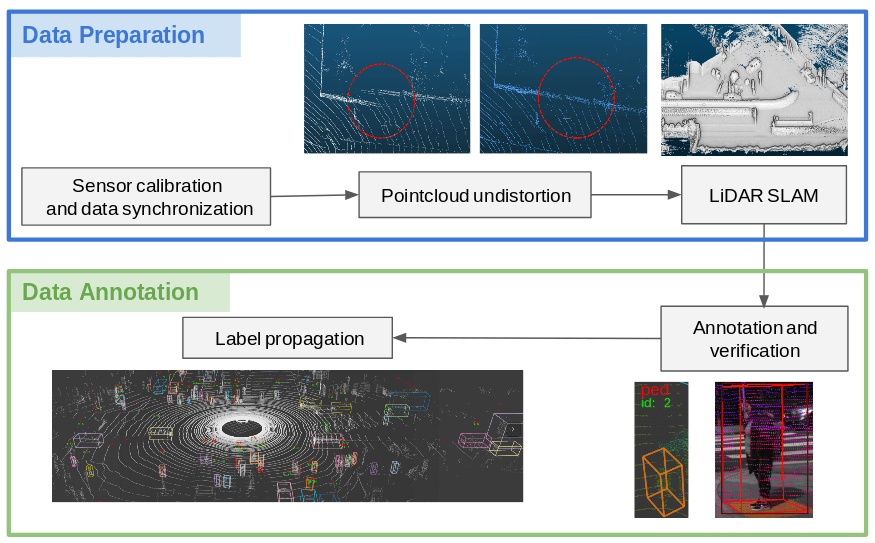}
    \caption{Data labelling procedure}
    \label{fig:flow_chart}
\end{figure*}
\section{H3D DATASET} \label{h3d}
An outline of the steps involved in dataset generation is shown in Fig.~\ref{fig:flow_chart}:
\begin{itemize}
\item Calibration between GPS/IMU (ADMA sensor) and LiDAR (Velodyne HDL-64E) is obtained using hand-eye calibration method~\cite{hand_eye} which is a well-known approach to find the relationship between two given trajectories from different coordinate system. Data from all five sensors (3 cameras, LiDAR and GPS/IMU) is time-synchronized with GPS time-stamps. 
\item Undistortion in point cloud data is performed to remove motion artifacts for superior annotation.
\item Point cloud registration is done to get ego-vehicle odometry estimates and point cloud data in each scenario is transformed to a fixed set of \textit{World} coordinates.  
\item Annotation of objects in point clouds is done in the \textit{World} coordinates by a group of annotators.
\item The labeled data (bounding boxes and point cloud) is then converted back to \textit{Velodyne} coordinates and changed to raw point cloud data.
\end{itemize}
\subsection{Sensor Setup}
The vehicle is equipped with the following sensors (as shown in Fig.~\ref{fig:vehicle}):
\begin{itemize}
\item three color PointGrey Grasshopper3 video cameras (30HZ frame rate, $1920\times1200$ resolution and $90\degree$ field-of-view (FOV) for left and right, $80\degree$ FOV for center)
\item a Velodyne HDL-64E S2 3D LiDAR (10 HZ  spin-rate, 64 laser beams, range: 100m, vertical FOV $26.9\degree$)
\item a GeneSys Eletronik GmbH Automotive Dynamic Motion Analyzer (ADMA) with DGPS output gyros, accelerometers and GPS (frequency: 100 HZ)
\end{itemize} 

Sensor data is recorded using a Ubuntu 14.04 machine with two eight-core Intel i5-6600K 3.5 GHz Quad-Core processors, 16 GB DDR3 memory, and a RAID 0 array of four 2TB SSDs.
%


\subsection{Data Collection}
\begin{figure}[h!]
\begin{center}
  \vbox{
      \subfigure[sensor layout of the vehicle \label{fig:vehicle_setup}]{\includegraphics[width=0.5\textwidth]{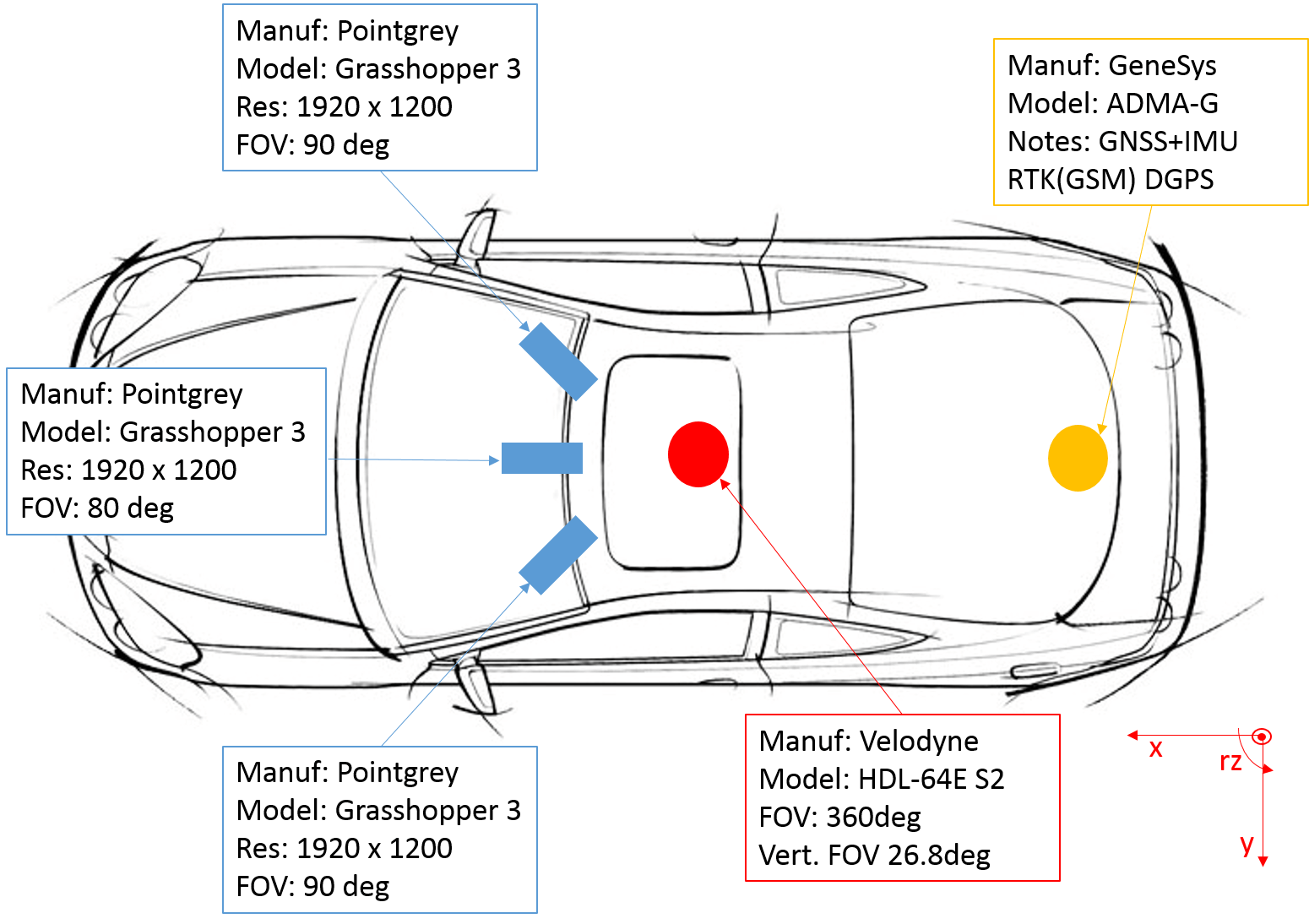}}\qquad
      \subfigure[side view of the vehicle \label{fig:vehicle_sideview}]{\includegraphics[width=0.5\textwidth]{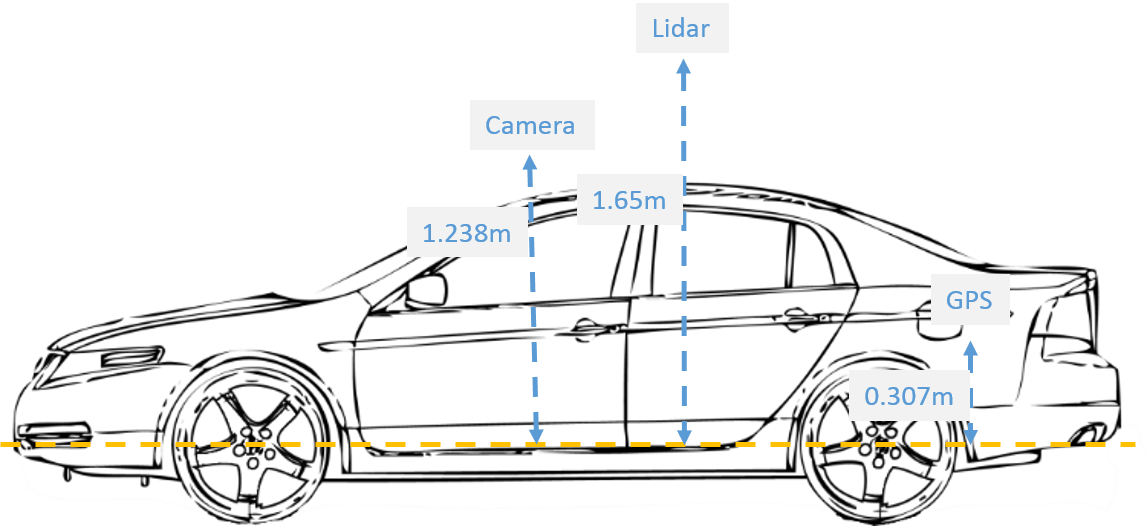}}
    }
\caption{Vehicle sensor setup}
\label{fig:vehicle}
\end{center}
\end{figure}

Data is collected in 4 urban areas in the San Francisco Bay Area from April to September 2017 using an instrumented vehicle shown in Fig.~\ref{fig:vehicle_setup}. 
The routes for data collection are overlaid on images from Google Earth as highlighted in Fig.~\ref{fig:data_region}. 
%
%
Sensor data is synchronized using Robot Operating System (ROS)\footnote{http://www.ros.org/} via a customized hardware setup.  
\subsection{Data Labelling Procedure}
In this section, we describe details of the 3D objects and tracklets labeling procedure for H3D.

\subsubsection{Data Preparation}
Cameras and LiDAR are hardware-timestamped using the GPS time-stamps and other sensor data is synchronized via ROS.
To prepare point cloud for annotation, an undistortion process is necessary because a raw point cloud is distorted due to a spinning LiDAR.
%
%

The process of undistortion is described as follows. 
%
%
The motion distortion is corrected using high-frequency fused GPS data obtained from the GPS/IMU sensor using linear interpolation method mentioned in~\cite{loam}.
%
%

%
\begin{figure}[h!]
    \centering
    \includegraphics[width=0.5\textwidth]{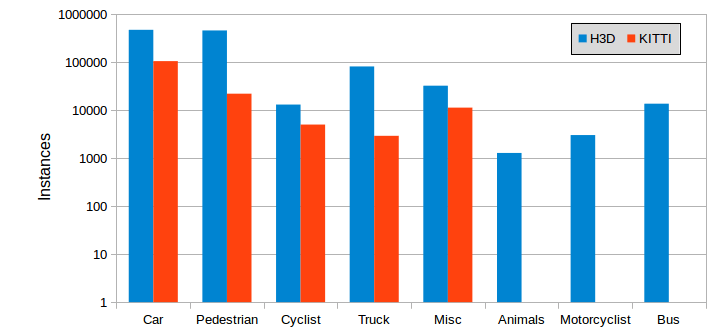}
    \caption{Distribution of classes in H3D and KITTI}
    \label{fig:data_distribution}
\end{figure}
\begin{figure*}[h!]
    \centering
    \includegraphics[width=0.8\textwidth]{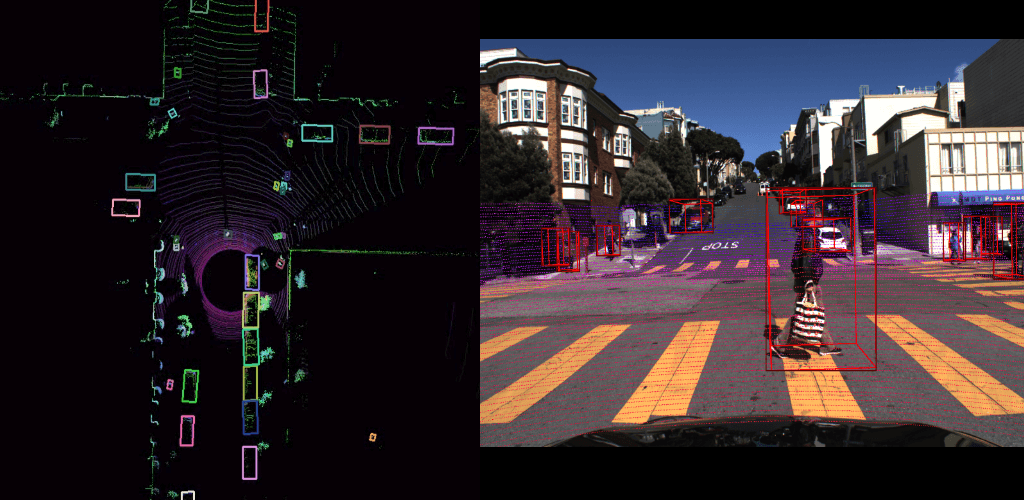}
    \caption{Data annotation verification by projecting annotation onto image and bird's eye view (BEV), with color coded track ID in BEV}
    \label{fig:data_verification}
\end{figure*}
Normal Distributive Transform (NDT)~\cite{ndt} method is used for point cloud registration as shown in Fig.~\ref{fig:flow_chart}. 
With each sequence being independent, point cloud is registered with respect to the initial frame (\textit{World}) of that particular sequence.
Such a registration process is needed for odometry estimation as GPS data is unreliable in urban areas with enclosed spaces and hence the transformation of point cloud to \textit{World} frame cannot be achieved accurately. 
Transforming point cloud data to \textit{World} coordinates simplifies the data annotation process as data association between static objects can be easily achieved given the correspondence between point cloud data in various frames.

\subsubsection{Data Annotation}\label{sec:data_annotation}
The registered point cloud data allows annotators to determine corresponding objects easily. 
Additionally, the three cameras are utilized to assist the annotation process in order to determine object categories.
We registered a sequence of point clouds at 2Hz from the odometry computed using NDT. 
Bounding boxes and track IDs are annotated on the registered point cloud.
Doing so, the static objects in the registered frames can be annotated in one shot and this significantly reduces the labeling efforts.
Moreover, the registered point cloud provides easier association of objects across frames.
The human-labeled annotations are then propagated to 10Hz using a linear interpolation technique, assuming a constant velocity model between each frame.
%
The labeled data (bounding boxes with track IDs) is transformed back to the \textit{Velodyne} coordinates using odometry estimates.
%
%

The quality of final labeled data is verified frame-by-frame by projecting the labeled bounding boxes onto corresponding images via methods similar to~\cite{calibration1,calibration2} and by visually inspecting the labeled data in BEV as shown in Fig.~\ref{fig:flow_chart}.

\subsection{Statistics}\label{statistics}
\begin{itemize}
\item \textbf{Complexity:} A comparison of density of common traffic participants averaged across  21 labeled scenarios in KITTI and 160 labeled scenarios in H3D is done to show the complexity of H3D dataset. For a fair comparison, number of annotations in H3D's 360$^\circ$ scene is assumed to be 4 times that of number of annotations in KITTI which are in frontal view of the scene. We observed that density of traffic participants in H3D is 15 times higher than that in KITTI.
\\
\item \textbf{Volume:} The total number of bounding box annotations and the various classes annotated are shown in Fig.~\ref{fig:data_distribution}. Also, it can be seen from Table~\ref{tab:num_instances} that the proportion of cars and pedestrians is consistent among training/validation/test datasets. 
\end{itemize}

\begin{table}
\begin{center}
\begin{tabularx}{0.5\textwidth}{X|X|X|X} 
  &\centering train &\centering validation &\centering\arraybackslash test \\ 
\hline
\centering Scenarios&\centering 50 &\centering 30 &\centering\arraybackslash 80 \\ 
\hline
\centering Frames&\centering 8873 &\centering 5170 &\centering\arraybackslash 13678 \\ 
\hline
\centering Car&\centering 157174 &\centering 84646 &\centering\arraybackslash 228738 \\ 
\hline
\centering Pedestrian &\centering 147985 &\centering 65549 &\centering\arraybackslash 242697 \\ 
\hline
\end{tabularx}
\caption{Instances for the train, validation, and test split}\label{tab:num_instances}
\end{center}
\end{table}
\section{3D DETECTION}
H3D is currently the only dataset that enables full 360-degree object detection in point cloud.
This paper evaluates VoxelNet~\cite{voxelnet} on H3D to obtain baseline values and assess the complexity of the dataset. 
%

A similar training procedure is adapted to that from the original literature (VoxelNet) with following modifications. 
Points within 40 meters radius of ego-vehicle are considered for car detection and points within 25.6 meters radius are considered for pedestrian detection.
The models for both car and pedestrian detection are trained using an ADAM optimizer. A learning rate of 0.01 is used for the first 40 epochs, then decreased to 0.001 for the next 20 epochs and further decreased to 0.0001 for the last 20 epochs (total 80 epochs). A batch size of 12 is used during training.
%

\begin{figure}[h!]
\begin{center}
  \mbox{
      \subfigure[all pedestrians and cars are detected with correct orientation \label{fig:successful_detection1}]{\includegraphics[width=0.23\textwidth]{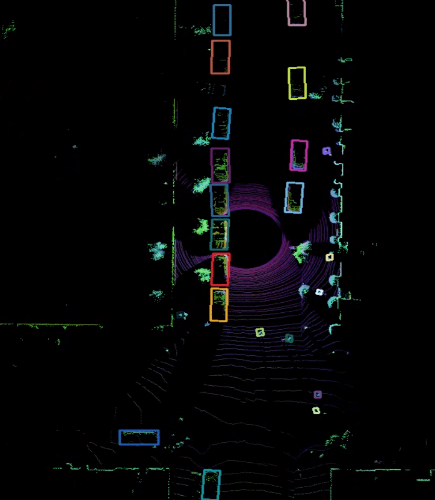}}\quad
      \subfigure[open area detections output \label{fig:successful_detection2}]{\includegraphics[width=0.24\textwidth]{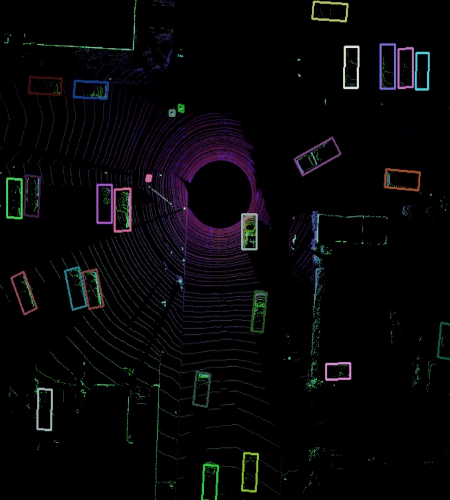}}
    }
  \caption{ Successful detection cases in BEV}
\label{fig:successful_detection}
\end{center}
\end{figure}

\begin{figure}[h!]
\begin{center}
  \mbox{
      \subfigure[wrong orientation for static vehicles in red dotted circle \label{fig:fail_detection1}]{\includegraphics[width=0.24\textwidth]{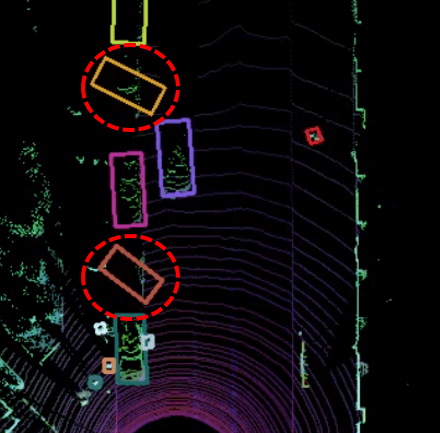}}\quad
      \subfigure[missing pedestrians in red dotted circle (because of occlusions) \label{fig:fail_detection2}]{\includegraphics[width=0.23\textwidth]{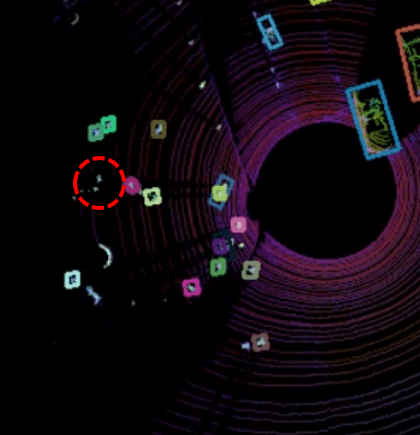}}
    }
  \caption{Failure detection cases in BEV}
  \label{fig:fail_detection}
\end{center}
\end{figure}

For evaluation, a similar protocol as KITTI is adapted. The IoU threshold for class car is set to 0.5 and that for class pedestrian is set to 0.25. 
Car and truck classes are combined when evaluating car detection performance. 
The results are summarized in (Table~\ref{tab:mAP_det}) and shown in Fig.~\ref{fig:successful_detection}.
The following challenges are encountered in 3D detection as highlighted in Fig.~\ref{fig:fail_detection}. The yaw estimation is not good where number of points for the particular object is less. Pedestrian detection fails to perform due to occlusion in crowded scenes.


\begin{table}
\begin{center}
\begin{tabularx}{0.5\textwidth}{X|X|X}
 \centering  &\centering BEV &\centering\arraybackslash 3D  \\ 
\hline
\centering car &\centering 76.50  &\centering\arraybackslash 68.31  \\  \hline
\centering pedestrian &\centering 50.88  &\centering\arraybackslash 50.39  \\ 
\hline
\end{tabularx}
\caption{mAP scores with 0.5 IoU for Car and 0.25 IoU for Pedestrian}\label{tab:mAP_det}
\end{center}
\end{table}
\section{3D Multi-Object Tracking}
3D objects are tracked using an Unscented Kalman Filter (UKF) via following four steps - prediction, data-association, update and track-management.
%
%
Data association of objects is done via euclidean distance between centroids of objects. 
%


\begin{figure*}[h!]
\begin{center}
  \mbox{
      \subfigure[frame number=1, starting frame in sequence \label{fig:fail_tracking1}]{\includegraphics[width=0.325\textwidth]{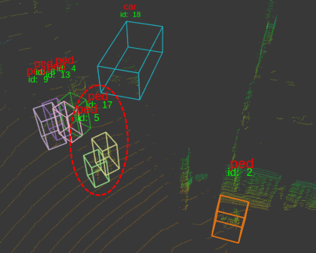}}
      \subfigure[frame number=22, pedestrians disappeared due to occlusion \label{fig:fail_tracking2}]{\includegraphics[width=0.325\textwidth]{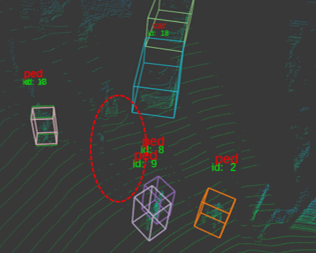}}
      \subfigure[frame number=34, new track IDs assigned to pedestrians \label{fig:fail_tracking3}]{\includegraphics[width=0.325\textwidth]{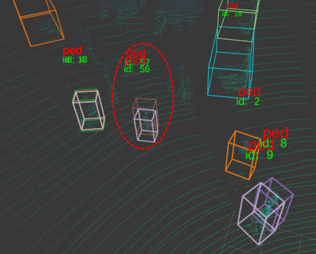}}
    }
  \caption{ Failure cases in long-term tracking via UKF for pedestrians; the red dotted circles in three different frames over time; (a) original pedestrian tracks; (b) highlight missing tracks due to occlusion; (c) change in data association}
\label{fig:fail_tracking}
\end{center}
\end{figure*}
Parameters used for tracking are summarized as follows. 
The state vector comprises of 5 variables, namely, 'x' and 'y' position of objects (in m), their velocities (in m/s), their orientation (in rad) and their angular velocities (in rad/s).
The euclidean distance threshold is set to 2 meters for data association for both car and pedestrian classes.  
An occlusion factor of 2, where occlusion factor is multiplied by the vertical area of object to determine if it becomes highly occluded.
Lastly, an aging factor of 2 is used such that an object is kept in the history of tracks for at most 2 frames.

The evaluation protocol from KITTI is adapted for tracking~\cite{kitti} with 0.5 3D IoU for both car and pedestrian classes. 
In the tracking algorithm evaluation, CLEAR MOT metrics are used~\cite{clear_mot} which include Multi-Object Tracking Precision (MOTP), Multi-Object Tracking Accuracy (MOTA) and Mostly Tracked (MT), Mostly Lost (ML) as mentioned in~\cite{mot_2}. 
The results for tracking are summarized in Table~\ref{tab:track} 
%
The analysis of tracking results shows that output is highly affected by quality of detections. 
The tracking algorithm is also evaluated with ground-truth locations of objects. The results indicate a considerable increase in accuracy with 0.99 MOTA, 1.00 MOTP, 1.00 MT and 0.00 ML for cars; 0.83 MOTA, 1.00 MOTP, 0.77 MT and 0.11 ML for pedestrians. 
Tracking is also affected when occlusions are present as shown in Fig.~\ref{fig:fail_tracking} for Pedestrians (track ID=15,17) with red dotted circle.  

\begin{table}
\begin{center}
\begin{tabularx}{0.5\textwidth}{ X|X|X|X|X} 
 \centering &\centering MOTA&\centering MOTP&\centering MT&\centering\arraybackslash ML\\  \hline
 \centering car&\centering  76.2 &\centering  73.1 &\centering 56.7&\centering\arraybackslash 25.1 \\ \hline
   \centering pedestrian&\centering 36.8 &\centering 63.0&\centering 14.1 &\centering\arraybackslash 43.4\\  \hline
\end{tabularx}
\caption{MOT scores with 0.5 3D IoU for Car and Pedestrian}\label{tab:track}
\end{center}
\end{table}
\section{CONCLUSION}
This paper demonstrates the uniqueness and importance of H3D for research on full-surround 3D multi-object detection and tracking in crowded urban scenes. 
%
Labelling methodology of H3D allows annotation of 3D objects and their track IDs on a large-scale efficiently.
%
%
A standard benchmark for future 3D point cloud detection and tracking algorithms development is established in the paper.
Given the significantly raised attention for 3D scene understanding, we
hope that H3D can push the performance envelope.
\\

\noindent \textbf{Acknowledgement:} We are grateful to our colleagues Behzad Dariush, Kalyani Polagani, Kenji Nakai, Athma Narayanan, and Wei Zhan for their valuable input.
\\ 
\bibliographystyle{IEEEtran}

\end{document}